%% file: ms.tex
\pdfoutput=1

\documentclass[10pt]{article}

\usepackage{Sty/mcr}
\usepackage{Sty/pkg}
\usepackage{Sty/thmE}

\usepackage{subcaption}

\usepackage{Sty/algorithm} 
\usepackage{Sty/algorithmic}

\usepackage{Sty/cancel}

\usepackage{Sty/soul}

\usepackage{hyperref}

\usepackage{Sty/bbm}

\usepackage[margin=1.2in]{Sty/geometry}
\usepackage{Sty/natbib}

\title{Statistical Inference for Clustering-based Anomaly Detection}

\date{\today}

\author{
  Nguyen Thi Minh Phu\textsuperscript{1,2}, 
  Duong Tan Loc\textsuperscript{1,2}, 
  Vo Nguyen Le Duy\textsuperscript{1,2,3}\thanks{Corresponding author: Email: duyvnl@uit.edu.vn}
}

\date{
\textsuperscript{1}University of Information Technology, Ho Chi Minh City, Vietnam\\
\textsuperscript{2}Vietnam National University, Ho Chi Minh City, Vietnam.\\
\textsuperscript{3}RIKEN, Japan.
}

\begin{document}

\maketitle

\begin{abstract}
\input{abst}

\end{abstract}

\clearpage


\input{sec1}
\input{sec2}
\input{sec3}
\input{sec4}

\input{sec5}




\bibliographystyle{abbrvnat}
\bibliography{ref}


\appendix

\input{appendix}

\clearpage

\end{document}

%% file: abst.tex
Unsupervised anomaly detection (AD) is a fundamental problem in machine learning and statistics. 
A popular approach to unsupervised AD is clustering-based detection.
However, this method lacks the ability to guarantee the reliability of the detected anomalies.
In this paper, we propose SI-CLAD (Statistical Inference for CLustering-based Anomaly Detection), a novel statistical framework for testing the clustering-based AD results.
The key strength of SI-CLAD lies in its ability to rigorously control the probability of falsely identifying anomalies, maintaining it below a pre-specified significance level $\alpha$ (e.g., $\alpha = 0.05$).
By analyzing the selection mechanism inherent in clustering-based AD and leveraging the Selective Inference (SI) framework, we prove that false detection control is attainable.
Moreover, we introduce a strategy to boost the true detection rate, enhancing the overall performance of SI-CLAD.
Extensive experiments on synthetic and real-world datasets provide strong empirical support for our theoretical findings, showcasing the superior performance of the proposed method.

%% file: sec1.tex
\section{Introduction}
\label{sec:introduction}

Anomaly detection (AD) is a fundamental problem in data analysis that involves identifying instances that deviate significantly from normal patterns. These anomalies, also referred to as outliers, can arise due to various factors such as system malfunctions or fraudulent activities. 
Detecting anomalies is crucial across diverse domains, including cybersecurity \citep{ten2011anomaly}, healthcare \citep{ukil2016iot}, finance \citep{ahmed2016survey}, and industrial monitoring \citep{carratu2023novel}, where early detection of unusual behavior can prevent potential failures or threats.

In many real-world scenarios where datasets have no pre-defined labels, unsupervised AD is widely used to detect the anomalies. 
It assumes that normal data points form dense regions in the data space, while anomalies appear in sparser areas. One popular approach to unsupervised AD is clustering-based detection \citep{aggarwal2017introduction, ccelik2011anomaly, syarif2012unsupervised, li2021clustering, ripan2021data}, where data points are grouped into clusters based on similarity. Anomalies are then identified as points that do not fit well into any cluster or are located far from cluster centroids.

A major concern in clustering-based AD is the risk of erroneous detection, where normal observations are mistakenly identified as anomalies. 
These errors, commonly referred to as \emph{false positives} (FPs), can have serious consequences, especially in high-stakes applications like fraud detection, medical diagnostics, and cybersecurity. 
%
%
For example, in fraud detection, frequent false alarms may result in legitimate transactions being blocked, causing frustration for users and financial losses for businesses. Similarly, in medical diagnostics, FPs could lead to unnecessary treatments or psychological distress for patients. 
Therefore, developing a statistical inference method to properly control the false positive rate (FPR) is critically needed.

However, achieving FPR control in statistical inference for clustering-based AD is challenging because both the detection and testing of anomalies are conducted on the same data.
In traditional statistics, selecting and testing a hypothesis on the same data is often referred to as \emph{double dipping} \citep{kriegeskorte2009circular}. It has been recognized that calculating a $p$-value for statistical inference in double dipping is highly biased, leading to a failure in properly controlling the FPR.

To address the challenge, our idea is to leverage  the \emph{Selective Inference (SI)}  framework introduced by \citet{lee2016exact}.
The key concept of Selective Inference (SI) is to perform statistical inference \emph{conditional on} the selected hypothesis, allowing us to avoid the selection bias that arises from double dipping.
In this paper, we utilize this concept to introduce a method that conducts statistical inference conditional on the AD results obtained from a clustering-based AD method.
With the proposed method, the FPR is properly controlled at a pre-specified level of guarantee $\alpha$ (e.g., 0.05), ensuring rigorous statistical reliability.

In recent years, several SI-related works have been proposed for the AD problem, including testing anomalies in linear regression \citep{chen2019valid, tsukurimichi2021conditional, phong2024controllable}, salient regions in deep learning models \citep{duy2022quantifying, daiki2023valid, shiraishi2024statistical}, and evaluating the  results detected by $k$-nearest neighbors AD \citep{niihori2025statistically}. However, none of the existing methods can be used to test anomalies detected by clustering-based AD. 
Developing an SI-based method depends on both the problem and the specific AD approach, making existing methods inapplicable to our setting. To achieve FPR control, we must carefully analyze the AD strategy of the clustering-based approach.
We would like to note that we start this research direction with the DBSCAN method \citep{ester1996density}, the most widely used robust clustering technique for AD, which has received substantial attention in both theory and practice. The detailed discussions on future extensions to other clustering-based AD techniques are provided in \S \ref{sec:sec5}.

\vspace{5pt}

\begin{figure*}[!t] 
\centering
    \includegraphics[width=\linewidth]{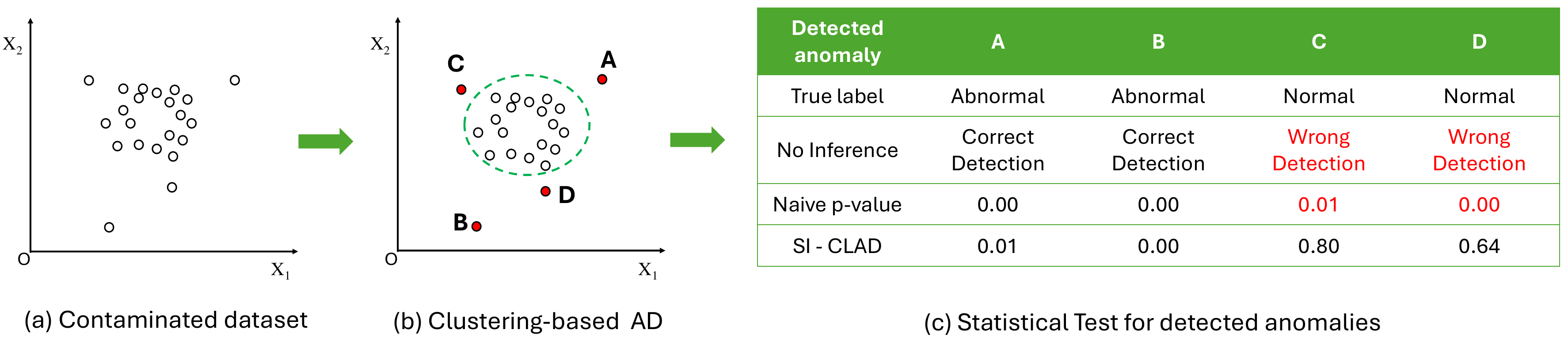}
    \caption{Illustration of the proposed SI-CLAD method. Performing clustering-based anomaly detection (AD) produces wrong anomalies (\textbf{C, D}). The naive $p$-values are even small for falsely detected anomalies. With the  
proposed SI-CLAD, we can identify both false positive (FP) and true positive (TP) detections, i.e., large \textit{p}-values for FPs and small \textit{p}-values for TPs.}
    \label{fig:illustration}
    \vspace{-10pt}
\end{figure*}

\textbf{Contributions.} First, we propose a novel statistical method, \emph{SI-CLAD} (Statistical Inference for CLustering-based AD), for clustering-based AD with controllable FPR. To the best of our knowledge, this is the first work capable of providing a valid $p$-value for testing the anomaly results obtained by clustering-based AD.
Second, we decompose the problem into multiple tractable sub-problems to enable an efficient test with the highest TPR while controlling the FPR. This approach leverages the divide-and-conquer principle.
Finally, we conduct experiments on both synthetic and real-world datasets to validate our theoretical results on successful FPR control and demonstrate the superior performance of the proposed SI-CLAD method compared the competitors.
Our implementation is available at 
\begin{center}
\href{https://github.com/ntmphu/SI-CLAD}{https://github.com/ntmphu/SI-CLAD}. 
\end{center}
The Python package can be installed via pip, i.e., pip install si-clad.
The illustration of the proposed SI-CLAD method is shown in Fig. \ref{fig:illustration}.

\vspace{5pt}

\textbf{Related works.}
While AD has been extensively studied \citep{aggarwal2017outlier}, the application of hypothesis testing to evaluate the AD results remains largely unexplored.
A key challenge in statistical inference for AD lies in ensuring the validity of $p$-values.
In particular, traditional (naive) $p$-values are often computed under the assumption that anomalies are fixed in advance.
However, since anomalies are typically identified by AD methods, this assumption is violated, resulting the naive $p$-values invalid.
In the context of supervised AD, data-splitting (DS) techniques can be used to perform valid statistical tests.
In contrast, clustering-based AD falls under unsupervised AD, making DS inapplicable.
To ensure valid statistical inference with traditional $p$-values, controlling the FPR requires multiple testing correction.
A common approach is the Bonferroni correction, where the adjustment factor scales exponentially with the number of instances, $n$, specifically reaching $2^n$.
However, this factor can grow prohibitively large unless $n$ is relatively small, resulting in overly conservative statistical inference.

SI offers a promising solution to the shortcomings of traditional inference methods.
Rather than relying on the overly conservative Bonferroni correction, which scales as $2^n$, SI mitigates the issue through conditional inference.
By conditioning on the specific set of identified anomalies, SI effectively reduces the correction factor to 1.
This fundamental idea of conditional SI was first introduced in the seminal work of \cite{lee2016exact}.
Since then, SI has been extensively studied and applied to a wide range of problems, including feature selection~\citep{fithian2014optimal, tibshirani2016exact, yang2016selective, sugiyama2021more, duy2021more}, change point detection~\citep{umezu2017selective, hyun2018post, duy2020computing, sugiyama2021valid, jewell2022testing}, clustering~\citep{lee2015evaluating, inoue2017post, gao2022selective}, and segmentation~\citep{tanizaki2020computing}.

There are several closely related studies on statistical testing for anomaly detection (AD).
\citet{chen2020valid} first proposed a method for testing the features of a linear model after removing anomalies, which inspired subsequent work on testing anomalies detected by LAD and Huber regressions \citep{tsukurimichi2022conditional} and RANSAC \citep{phong2024controllable}.
SI has also been explored in testing the salient region detected by deep learning modesl \citep{duy2022quantifying, daiki2023valid, shiraishi2024statistical} and evaluating the anomalies in a semi-supervised setting \citep{niihori2025statistically}.
However, none of the existing methods can be directly applied to the clustering-based AD setting, which is the primary focus of this paper.

%% file: sec2.tex
\section{Problem Statement}
\label{sec:problem_statement}
To formulate the problem, let us consider a random vector:
\begin{equation} 
	\bm X = \left (X_1, ..., X_n \right)^\top = \bm \mu + \bm \veps, \quad \bm \veps \sim \NN(\bm 0, \Sigma), 
\end{equation}
where $n$ is the number of observations, $\bm \mu \in \RR^{n} $ is an unknown mean vector, $\bm \veps$ is the Gaussian noise vector with the covariance matrice $\Sigma \in \RR^{n \times n}$ assumed to be known or estimable from independent data. 

\subsection{DBSCAN-based Anomaly Detection}

Given a data $\bm X$, the goal of applying DBSCAN on $\bm X$ is to identify the clusters as dense regions in the data space and detect the set of anomalies.
The key idea is that, for each point of a cluster, its $eps$-neighborhood for a given radius of the neighborhood around a data point $eps > 0$ has to contain at least a minimum number of points $(MinPts)$.
The $eps$-neighborhood of a point $X_{i} $, $i \in [n] = \{1, 2, ...,n \} $, denoted by 
\[
N_{ eps}( X_{i}) = 
\left\{ 
X_{j} \in \bm X \mid \big \|X_i - X_j \big \|^2 \leq eps^2
\right \}.
\]
%

There are two kinds of points in a cluster: 

\vspace{5pt}
{\noindent $\bullet$} \textbf{Core points:} $X_i \in \bm X$ is a core point if $\big |N_{eps}(X_i) \big | \geq MinPts$.

{\noindent $\bullet$} \textbf{Border points:} $X_i \in \bm X$ is a border point  is a border point if it lies within the $eps$-neighborhood of a core point but does not satisfy the core point condition.

\vspace{5pt}
The procedure of detecting anomalies using DBSCAN \citep{ester1996density} is summarized as follows:

\vspace{5pt}
{\noindent $\bullet$} \textbf{Identifying core points}: For each $X_i$, for any $i \in [n]$, determine whether it qualifies as a core point.


{\noindent $\bullet$} \textbf{Forming and expanding clusters}: For each core point not yet assigned to a cluster, create a new cluster.
Recursively add all density-reachable points (both core and border) by iteratively querying the $eps$-neighborhood of core points in the cluster until no further expansion is possible.


{\noindent $\bullet$} \textbf{Detecting anomalies}: After processing all points, any data point that does not belong to a cluster is determined as an anomaly.
We define the function that maps the data $\bm X$ to the set $\cO$ of indices of anomalies as
\begin{align} \label{eq:detected_anomalies}
	\cA: \bm X \mapsto \cO \subseteq [n],
\end{align}
where $\cA$ indicates the DBSCAN algorithm.

\subsection{Statistical Inference on the Detected Anomalies}

In this section, we formulate the problem  of testing if the anomalies in \eq{eq:detected_anomalies} are truly abnormal. 

\textbf{The null and alternative hypotheses.} To statistically quantifying the significance of the detected anomalies, we consider the following statistical test:
\begin{align} 
	{\rm H}_{0, j}: \mu_j = \bar{\bm \mu}_{- \cO}
	\quad
	\text{vs.}
	\quad 
	{\rm H}_{1, j}: \mu_j \neq \bar{\bm \mu}_{- \cO}, \quad 
	\forall j \in \cO,
\end{align}
where ${\bm \mu}_{- \cO}$ represents the mean of true values of the remaining data points after excluding the set of detected anomalies  ${\cO}$:
\[
	\bar{\bm \mu}_{- \cO} = 
	\frac{1}{n - |\cO|} \sum \limits_{\ell \in [n] \setminus \cO}
	\mu_\ell .
\]

In essence, our goal is to determine whether each detected anomaly $j \in \cO$ truly deviates from the remaining data points after excluding  $\cO$. The null hypothesis \( {\rm H}_{0, j} \) posits that the true value \( \mu_j \) is equal to the mean of the remaining true values, implying that the \( j^{th} \) point is \emph{not} a true anomaly. In contrast, the alternative hypothesis \( {\rm H}_{1, j} \) asserts that \( \mu_j \) significantly deviates from \( \bar{\bm \mu}_{-\cO} \), indicating that the \( j^{th} \) point is a true anomaly.

\textbf{Test statistic.} 
To test the above hypotheses, the test statistic is defined as:
\begin{equation}\label{eq:test_statistic}
    T_j = X_j - \bar{\bm X}_{-\cO} = \bm \eta_j^\top \bm X, 
\end{equation}
where 
%
$
    \bm \eta_j = 
    \bm e_j - \frac{1}{n - |\cO|} \bm e_{-\cO} \in \RR^n
$
%
is the direction of the test statistic, \( \bm e_j \) is a vector in \( \mathbb{R}^n \) with the \( j^{\mathrm{th}} \) entry equal to \( 1 \) and all other entries equal to \( 0 \), while \( \bm e_{-\cO} \) is a vector in \( \mathbb{R}^n \) with the \( j^{\mathrm{th}} \) entry equal to \( 0 \) for all \( j \in \cO \) and \( 1 \) otherwise.
After computing the test statistic in \eq{eq:test_statistic}, we calculate the corresponding $p$-value. 
Given a significance level $\alpha \in [0, 1]$, e.g., 0.05, we reject the null hypothesis ${\rm H}_{0, j}$ and assert conclude that $X_j$ is an anomaly if the $p$-value less than or equal to $\alpha$. Conversely, if the $p$-value is greater than $\alpha$, we infer that there is insufficient evidence to conclude that $X_j$ is an anomaly.

\textbf{The invalidity of traditional 
$p$-value computation.}
Let $\bm X^{\rm obs}$ be an observation (realization) of the random vector $\bm X$. Suppose that the hypotheses in \eq{eq:test_statistic} are fixed, i.e., non-random, the traditional (naive) $p$-value is computed as:
\begin{align}
	p^{\rm naive}_j = 
	\mathbb{P}_{\rm H_{0, j}} 
	\left ( 
		\left | \bm \eta_j^\top \bm X \right |
		\geq 
		\left | \bm \eta_j^\top \bm X^{\rm obs} \right |
	\right ), 
\end{align}
If the vector $\bm \eta_j$ in \eq{eq:test_statistic} is independent of the data, the naive $p$-value remains valid in the sense that
\begin{align} \label{eq:valid_p_value}
	\mathbb{P} \Big (
	\underbrace{p_j^{\rm naive} \leq \alpha \mid {\rm H}_{0, j} \text{ is true }}_{\text{a false positive}}
	\Big) = \alpha, ~~ \forall \alpha \in [0, 1],
\end{align} 
i.e., the probability of obtaining a false positive is controlled under a certain level of guarantee.
However, in our setting, the vector $\bm \eta_j$ actually depends on the set of anomalies $\cO$, which is obtained by applying DBSCAN to the data,  leading to \emph{selection bias} in the analysis.
As a result, the validity condition of the $p$-value in \eq{eq:valid_p_value} is no longer satisfied, making the naive $p$-value \emph{invalid}.

%% file: sec3.tex
\section{Proposed Method}
\label{sec:sec3}

In this section, we detail the proposed SI-CLAD method.
We establish valid $p$-values for anomalies detected by DBSCAN by leveraging the SI framework \citep{lee2016exact}.
Furthermore, we introduce an efficient approach for computing these $p$-values based on the concept of divide-and-conquer.
\begin{figure*}[!t] 
\centering
    \includegraphics[width=\linewidth]{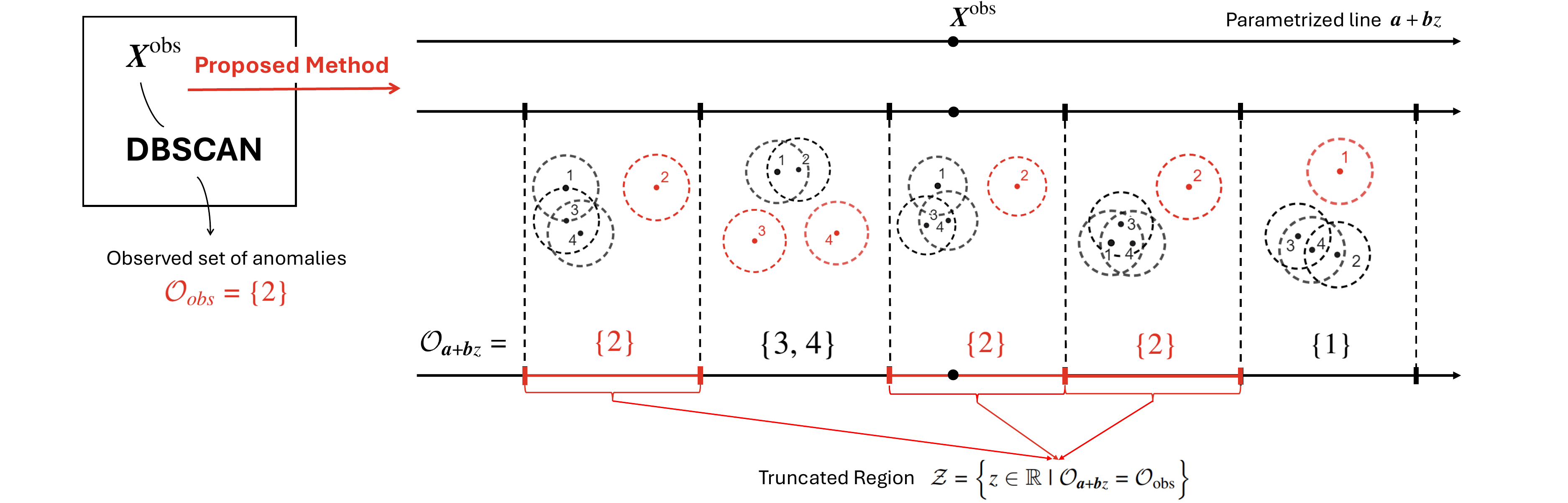}
    \caption{A schematic illustration of the proposed method. By applying DBSCAN to the observed data, we obtain a set of anomalies. Then, we parametrize the data with a scalar parameter $z$ in the dimension of the test statistic to identify the truncation region $\mathcal{Z}$ whose data have the same result of anomaly detection as the observed data. Finally, the valid inference is conducted conditional on $\mathcal{Z}$. We utilize the concept of divide-and-conquer and introduce a hierarchical line search method for efficiently characterizing the truncation region $\mathcal{Z}$}
    \label{fig:illustration_methodology}
    \vspace{-10pt}
\end{figure*}

\vspace{5pt}
\subsection[The Valid p-value in SI-CLAD]{%
  The Valid \texorpdfstring{$p$}{p}\nobreakdash-value in SI-CLAD%
}
\label{subsec:valid_pvalue_SI-CLAD}

Leveraging the concept of SI, we need to conduct \emph{conditional} inference to eliminate the information used in the initial hypothesis generation process, thereby avoiding selection bias.
In our setting, we specifically consider the sampling distribution of the test statistic \emph{conditional} on the selection event that DBSCAN identifies the set of anomalies $\cO$. Thus, we introduce the \emph{selective $p$-value} defined as:
\begin{align} \label{eq:selective_p}
	p^{\rm selective}_j = 
	\mathbb{P}_{\rm H_{0, j}} 
	\Big ( 
		\left | \bm \eta_j^\top \bm X \right |
		\geq 
		\left | \bm \eta_j^\top \bm X_{\rm obs} \right |
		~
		\Big | 
		~
	\cO_{\bm X}
	=
	\cO_{\rm obs},
	\cQ_{\bm X}
	=
	\cQ_{\rm obs}
	\Big ), 
\end{align}
where $\cO_{\bm X} = \cO_{\rm obs}$ denotes the event that the anomalies detected for the random vector $\bm X$ are the same as those for the observed data  $\bm X^{\rm obs}$. The second condition $\cQ_{\bm X}$ is the \emph{nuisance component} is given by 
\begin{align} \label{eq:q_and_b}
	\cQ_{\bm X} = 
	\left ( 
	I_n - 
	\bm b
	\bm \eta_j^\top \right ) 
	{\bm X}, \text{ where } \bm b = \frac{\Sigma \bm \eta_j}
	{\bm \eta_j^\top \Sigma \bm \eta_j}
\end{align}

\begin{remark}
The nuisance component $\cQ_{\bm X}$ corresponds to
the component $\bm z$ in the seminal paper of \cite{lee2016exact} (see Sec. 5, Eq. (5.2), and Theorem 5.2)).
We note that additionally conditioning on $\cQ_{\bm X}$, which is required for technical purpose, is a standard approach in the SI literature and it is used in almost all the SI-related works that we cited.
%
\end{remark}

\begin{lemma} \label{lemma:valid_selective_p}
The selective $p$-value proposed in \eq{eq:selective_p} satisfies the property of a valid $p$-value:
\begin{align*}
	\mathbb{P}_{\rm H_{0, j}}  \Big (
	p_j^{\rm selective} \leq \alpha
	\Big) = \alpha, ~~ \forall \alpha \in [0, 1].
\end{align*} 
\end{lemma}

\begin{proof}
The proof is deferred to the supporting information (A. APPENDIX).
\end{proof}
Lemma \ref{lemma:valid_selective_p} states that using the selective $p$-value ensures theoretical control of the FPR at the significance level $\alpha$.
The computation of the selective $p$-value in $\ref{eq:selective_p}$ requires characterizing the conditioning event $\left \{ \cO_{\bm X} = \cO_{\rm obs}, \cQ_{\bm X} = \cQ_{\rm obs} \right\}$, which will be discussed in the next section.

\subsection{Characterization of the Conditioning Event}
\label{subsec:conditional_data_space}

We define the conditional data space of $\bm X  \in \RR^n$ that satisfies the conditions in \eq{eq:selective_p} as:
\begin{align} \label{eq:conditional_data_space}
	\cX = \left \{ 
	\bm X \in \RR^n \Big | ~
	\cO_{\bm X}
	=
	\cO_{\rm obs}, 
	\cQ_{\bm X}
	=
	\cQ_{\rm obs}
	\right \}. 
\end{align}
In fact, as stated in the following lemma, $\cX$ is restricted to \emph{a line} in $\RR^n$.
\begin{lemma} \label{lemma:data_line}
The set $\cX$ in \eq{eq:conditional_data_space} can be rewritten using a scalar parameter $z \in \RR$ as follows:
\begin{align} \label{eq:conditional_data_space_line}
	\cX = \big \{ \bm X (z) = \bm a + \bm b z \mid z \in \cZ \big \},
\end{align}
where vector $\bm a = \cQ_{\rm obs}$, $\bm b$ is defined in \eq{eq:q_and_b}, and
\begin{align} \label{eq:cZ}
	\cZ = \Big \{ 
	z \in \RR 
	\mid 
	\cO_{\bm X(z)} = \cO_{\rm obs}
	\Big \}.
\end{align}

\end{lemma}
\begin{proof}
The proof is deferred to the supporting information (A. APPENDIX).
\end{proof}

Let $Z \in \RR$ be a random variable, and let $Z^{\rm obs}$ be its observed value, satisfying $\bm X = \bm a + \bm b Z$ and $\bm X^{\rm obs} = \bm a + \bm b Z^{\rm obs}$. Then, the selective $p$-value in (\ref{eq:selective_p}) can be rewritten as:
\begin{align} \label{eq:selective_p_reformulated}
	p^{\rm selective}_j = \mathbb{P}_{{\rm H}_{0, j}} \left ( \left | Z \right | \geq \left | Z^{\rm obs} \right |  \big | ~    Z \in \cZ \right ).
\end{align}
Since $Z \sim \NN(0, \bm \eta_j^\top \Sigma \bm \eta_j)$ under ${\rm H}_{0, j}$, $Z \mid Z \in \cZ$ follows a \emph{truncated} normal distribution. 
Once the truncation region $\cZ$ in \eq{eq:cZ} is determined, computation of the selective $p$-value in (\ref{eq:selective_p_reformulated}) becomes straightforward.
Thus, the key remaining task is to identify $\cZ$.

\subsection[
  Identification of the Truncation Region Z
]{%
  Identification of the Truncation Region \texorpdfstring{$\cZ$}{Z}%
}
\label{subsec:identification_cZ}

Due to the inherent complexity of the DBSCAN algorithm, directly identifying $\cZ$ is intrinsically challenging. To mitigate this, it is common practice in the SI literature to introduce additional conditioning events that occur during the execution of the algorithm. This additional conditioning, often referred to as \textit{over-conditioning}, enhances computational tractability but is not necessary for ensuring valid inference. While $p$-values computed under over-conditioning remain valid, they often result in a substantial reduction in the statistical power of the test. To address this limitation, we propose an efficient approach (demonstrated in Fig. \ref{fig:illustration_methodology}) to identify $\cZ$ as follows:

\begin{itemize}
    \item We define the over-conditioning problem by additionally conditioning on the $eps$-neighborhood points of all data points.
    
    \item We prove that the over-conditioning problem is effectively solvable.
        
    \item The truncation region $\cZ$ in \eq{eq:cZ} is constructed by combining multiple over-conditioning problems along the parameterized line defined in \eq{eq:conditional_data_space_line}.
    
\end{itemize}

%
%

\textbf{Over-conditioning on $eps$-neighborhood points.} Consider a data $\bm X^\prime \in \RR^n$, we define the over-conditioning region as follows:
 \begin{align} \label{eq:z_oc}
\cZ^{\rm oc}\big (\bm X^\prime \big ) = \bigcap \limits_{i \in [n]} \Big \{ z \in \RR \mid N_{eps}\big (X_i(z) \big ) = N_{eps}(X^{\prime}_i) \Big \}.
\end{align}
The computation of $\cZ^{\rm oc}\big (\bm X^\prime \big )$ is described in the following lemma.

\begin{lemma} \label{lemma:over_conditioning}
The over-conditioned region \(\mathcal{Z}^{oc}\big (\bm X^\prime \big )\) defined in Equation \eqref{eq:z_oc} can be characterized by a set
of quadratic inequalities w.r.t. z described as follows:  
\begin{align*}
    \cZ^{\rm oc}\big (\bm X^\prime \big ) =  \left\{ z \in \mathbb{R} \;\middle|\; \bm p z^2 + \bm q z + \bm t \leq 0 \right\},
\end{align*}
where vectors $\bm p$, $\bm q$, $\bm t$ are defined in the supporting information (A. APPENDIX).
\end{lemma}

\begin{proof}
The proof is deferred to the supporting information (A. APPENDIX).
\end{proof}

\textbf{Computation of truncated region $\cZ$ via Parametric Programming \citep{duy2021more}.}
Once the over-conditioned region along the parametric line $\bm a + \bm bz$ has been computed, we need to consider the union of them to get the minimum-conditioned region:
\begin{align} \label{eq:union_Zoc}
    \cZ = \bigcup \limits_{z \in \RR ~ \mid ~ \mathcal{O}_{\bm X(z)} = \mathcal{O}_{\rm obs}} \cZ^{\rm oc}\big (\bm X(z) \big )
\end{align}

Our approach is to identify $\cZ$ by repeatedly applying DBSCAN w.r.t. $MinPts$ and $eps$ to a sequence of datasets $\bm a + \bm b z$ within sufficently wide range of $z \in \left[ z_{min}, z_{max} \right]$. From the above discussion, ${Z}^{\rm oc} \big ( \bm X^\prime\big )$ is given by the intersection of a finite number of quadratic inequalities. Thus, solving them analytically, $\cZ$ can be represented as a union of multiple non-overlapping intervals.
%
The procedure of computing the truncated region can be summarized in Algorithm \ref{alg:compute_solution_path}. Subsequently, it is straightforward to calculate the proposed selective inference through Eq. \ref{eq:selective_p_reformulated}. The entire procedure of the proposed SI-CLAD method is shown in Algorithm \ref{alg:SI_detected_outliers}. 

\begin{algorithm}[!t]
\renewcommand{\algorithmicrequire}{\textbf{Input:}}
\renewcommand{\algorithmicensure}{\textbf{Output:}}
\begin{algorithmic}[1]
 \REQUIRE $\bm a, \bm b, z_{\rm min}, z_{\rm max}, MinPts, eps, \mathcal{O}_{\rm obs}$
	\vspace{4pt}
	\STATE Initialization: $z  = z_{\rm min}$, $\cZ = \emptyset$
	\vspace{4pt}
	\WHILE {$z < z_{\rm max}$}
		\vspace{4pt}
        \STATE $\bm X (z) = \bm a + \bm b z$
        \vspace{4pt}
       \STATE Compute $N_{eps}\big (\bm X(z) \big )$ 
		\vspace{4pt}
		\STATE $\cO_{\bm X(z)} \leftarrow$ DBSCAN on $\bm X^\prime(z)$ with  $MinPts$ and $eps$
		\vspace{4pt}
		\STATE Compute $ [L^{z}, R^{z}] = \cZ^{\rm oc} \big (\bm X(z) \big )\leftarrow$ Lemma \ref{lemma:over_conditioning} 
		\vspace{4pt}
        
       \IF {$\cO_{\bm X(z)} = \cO_{\rm obs}$}
		\vspace{4pt}
		\STATE $\cZ \leftarrow \cZ \cup [L^{z}, R^{z}] $
		\vspace{4pt}
		\ENDIF
		\vspace{4pt}
       \STATE $z \leftarrow R^{z} + \delta$ \COMMENT{Step past $R^{z}$ by tolerance $\delta$ (e.g., 0.001)}
        
	\ENDWHILE
	\vspace{4pt}
 \ENSURE $\cZ$ 
\end{algorithmic}
\caption{\tt parametrized\_line\_search}
\label{alg:compute_solution_path}
\end{algorithm}

\begin{algorithm}[!t]
\renewcommand{\algorithmicrequire}{\textbf{Input:}}
\renewcommand{\algorithmicensure}{\textbf{Output:}}
\begin{algorithmic}[1]
 \REQUIRE $\bm X^{\rm obs}, MinPts, eps, z_{\rm min}, z_{\rm max}$
	\vspace{4pt}
	\STATE $\cO_{\rm obs} \leftarrow$ DBSCAN on $\bm X^{\rm obs}$ with  $MinPts$ and $eps$
	\vspace{4pt}
	\FOR {$j \in \cO_{\rm obs}$}
		\vspace{4pt}
		\STATE Compute $\bm \eta_j \leftarrow$ Eq. \eq{eq:test_statistic}, $\bm a$ and $\bm b \leftarrow$ Eq. \eq{eq:conditional_data_space_line}
		\vspace{4pt}
		\STATE  $ \cZ \leftarrow$ {\tt parametrized\_line\_search} \big ($\bm a, \bm b, z_{\rm min}, z_{\rm max}, MinPts, eps, \mathcal{O}_{ \rm obs} \big )$
		\vspace{4pt}
		\STATE $p^{\rm selective}_j \leftarrow$  Eq. \eq{eq:selective_p_reformulated} 
		\vspace{4pt}
	\ENDFOR
	\vspace{4pt}
 \ENSURE $\big \{ p^{\rm selective}_j \big \}_{j \in \cO_{\rm obs}}$ 
\end{algorithmic}
\caption{{\tt SI\_CLAD}}
\label{alg:SI_detected_outliers}
\end{algorithm}

\subsection{Extension to Multi-Dimension}\label{subsec:extension}

In this section, we extend the problem setup and the proposed method in multi-dimensional data. We consider a random matrix \( X \in \mathbb{R}^{n \times d} \), where \( n \) is the number of observations and \( d \) is the dimensionality of each observation. $X$ is a random sample from
\[
\mathrm{vec}(X)\;\sim\;\mathbb{N}\bigl(\mathrm{vec}(\mathbf{M}),\,\bm{\Sigma}_{\mathrm{vec}(X)}\bigr)
\]
%
%
where the operator \( \mathrm{vec}(.) \) converts a matrix into a vector by stacking the columns of the matrix on top of each other, $\mathbf{M}$ is an unknown signal matrix, and $\bm{\Sigma}_{\mathrm{vec}(X)}$ is the covariance matrix assumed to be known or estimable from independent data.
The result after applying DBSCAN w.r.t. $eps$ and $MinPts$ to $X$ is:
\begin{align*}
	\cA: X \mapsto \cO \in [n].
\end{align*}
for $j \in \cO$, we consider the following hypotheses:
\begin{align*}
	{\rm H}_{0, j}:  M_{j, \kappa} = \bar{ M}_{- \cO, \kappa}
	\quad
	\text{vs.}
	\quad 
	{\rm H}_{1, j}:  M_{j, \kappa} \neq \bar{M}_{- \cO, \kappa},
\end{align*}
for all $\kappa \in [d]$ and $\bar{M}_{- \cO, \kappa}$ is given by:
$
	\bar{M}_{- \cO, \kappa} = 
	\frac{1}{n - |\cO|} \sum \limits_{\ell \in [n] \setminus \cO}
	M_{\ell, \kappa}.
$
%
To test the hypotheses, the test statistic is defined as:
\begin{align} \label{eq:test_statistic_mul}
\Gamma_j = \frac{1}{d}\sum_{\kappa \in [d]} \left| X_{j, \kappa} - \bar{X}_{-\mathcal{O}, \kappa} \right| = \bm{\eta}_j^\top \mathrm{vec}(X),
\end{align}
with \( \bm{\eta}_j \in \mathbb{R}^{nd} \)  given by:
\[
\bm{\eta}_j = \frac{1}{d}\left( \mathbf{I}_d \otimes \bm{e}_j - \frac{1}{n - |\mathcal{O}|} \mathbf{I}_d \otimes \bm{e}_{-\mathcal{O}} \right) \bm{s},
\]
where $\mathbf{I}_d \in \mathbb{R}^{d\times d}$ denotes the $d$-dimensional identity matrix and $\bm e_j \in \mathbb{R}^n$ is the $j$-th standard basis vector. The vector $\bm e_{-\mathcal{O}} \in \mathbb{R}^n$ selects all coordinates except those indexed by the outlier set $\mathcal{O}$. The sign vector $\bm s = [s_1, s_2, \dots, s_d]^\top \in \mathbb{R}^d$ has entries $s_\kappa = \mathrm{sign}(X_{j,\kappa} - \bar X_{-\mathcal{O},\kappa})$, and the operator $\otimes$ denotes the Kronecker product.
The sign function \( \mathrm{sign}(x) \) is defined as:
\[
\mathrm{sign}(x) =
\begin{cases}
   \frac{x}{|x|}, & \text{if } x \neq 0, \\
   0, & \text{if } x = 0.
\end{cases}
\]
\newcommand{\xvec}{\mathbf{x}_{\mathrm{vec}}}
In order to compute the valid $p$-values for detected outliers, 
we consider the following distribution of the test statistic:
\begin{align} \label{eq:conditional_distribution_mul}
	\mathbb{P} \Big ( 
	\Gamma_j
	\mid
	\cO_{X}
	=
	\cO_{\rm obs}, ~
	{\bm s}_{X} 
	=
	{\bm s}_{\rm obs}
	\Big ).
\end{align}

By leveraging the technique in \S \ref{subsec:conditional_data_space} and \S \ref{subsec:identification_cZ} into the setting of multi-dimension, 
it is straightforward to obtain the conditional space in \ref{eq:conditional_distribution_mul} which is subsequently used to compute the $p$-value.

%% file: sec4.tex
\section{Experiments}
\label{sec:sec4}

In this section, we present the performace of the proposed method. We compared the result of the following methods in terms of FPR and TPR:
\begin{itemize}
    \item  {\tt SI-CLAD:} the proposed method
    
    \item  {\tt SI-CLAD-oc:} the proposed method with only the over-conditioning described in 
    \S\ref{subsec:identification_cZ}
    
    \item  {\tt Naive:} traditional statistical inference
    
    \item  {\tt Bonferroni:} the most popular multiple testing
    \item  {\tt No-Inference:} Cluster-based Anomaly Detection without inference
\end{itemize}

We note that if a method cannot control the FPR under $\alpha$, it is invalid and its TPR does not need to be considered. Additionally, a method with high TPR inherently corresponds to a low FNR.

\subsection{Numerical Experiments.}

\begin{figure}[!t]
\begin{minipage}{.48\linewidth}
\begin{subfigure}{\linewidth}
  \centering
  \includegraphics[width=\linewidth]{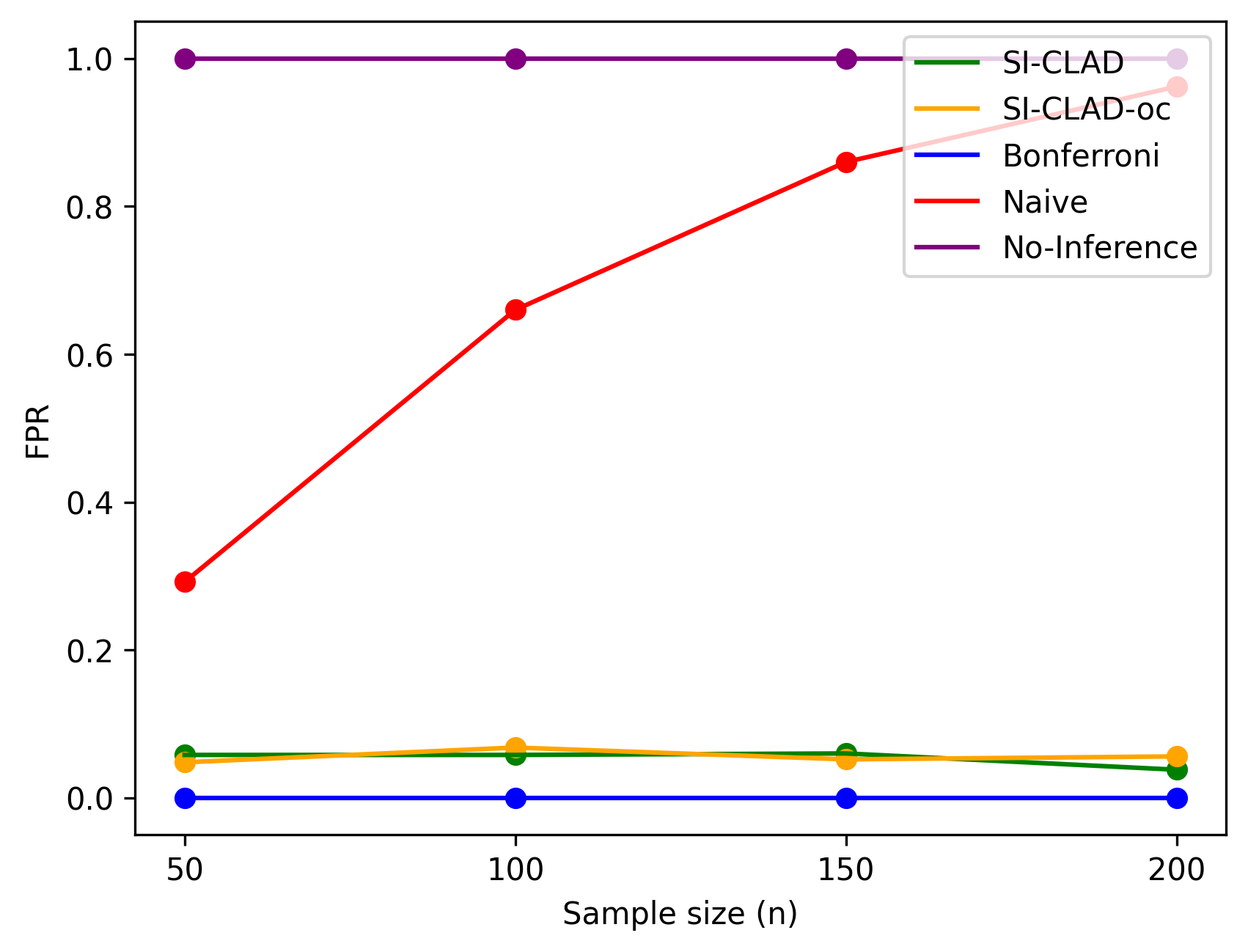}  
\end{subfigure}
\begin{subfigure}{\linewidth}
  \centering
  \includegraphics[width=\linewidth]{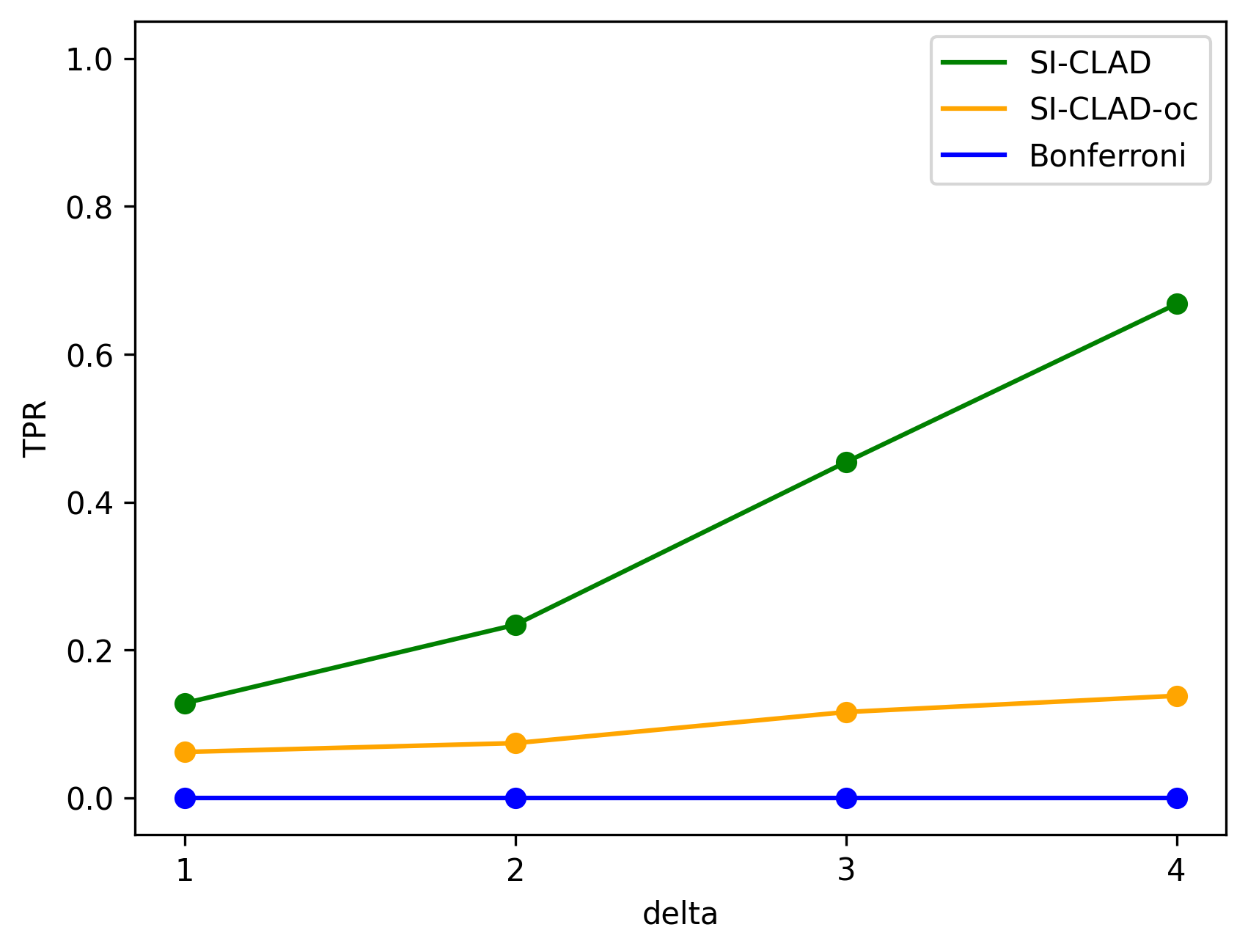} 
\end{subfigure}
\caption{FPR and TPR in univariate case} 
\label{fig:1d}
\end{minipage}
\hspace{10pt}
\begin{minipage}{.48\linewidth}
\begin{subfigure}{\linewidth}
  \centering
  \includegraphics[width=\linewidth]{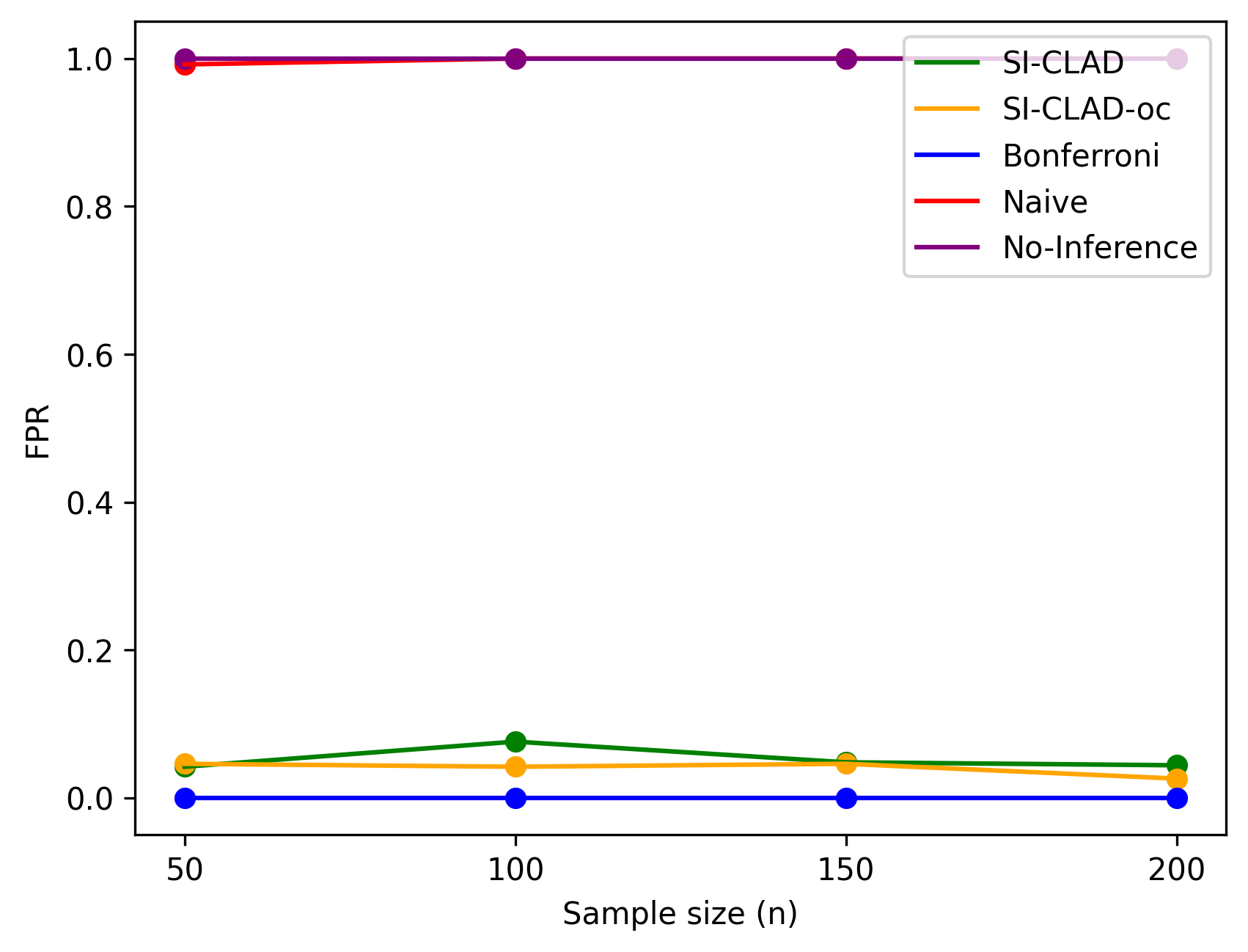}  
\end{subfigure}
\begin{subfigure}{\linewidth}
  \centering
  \includegraphics[width=\linewidth]{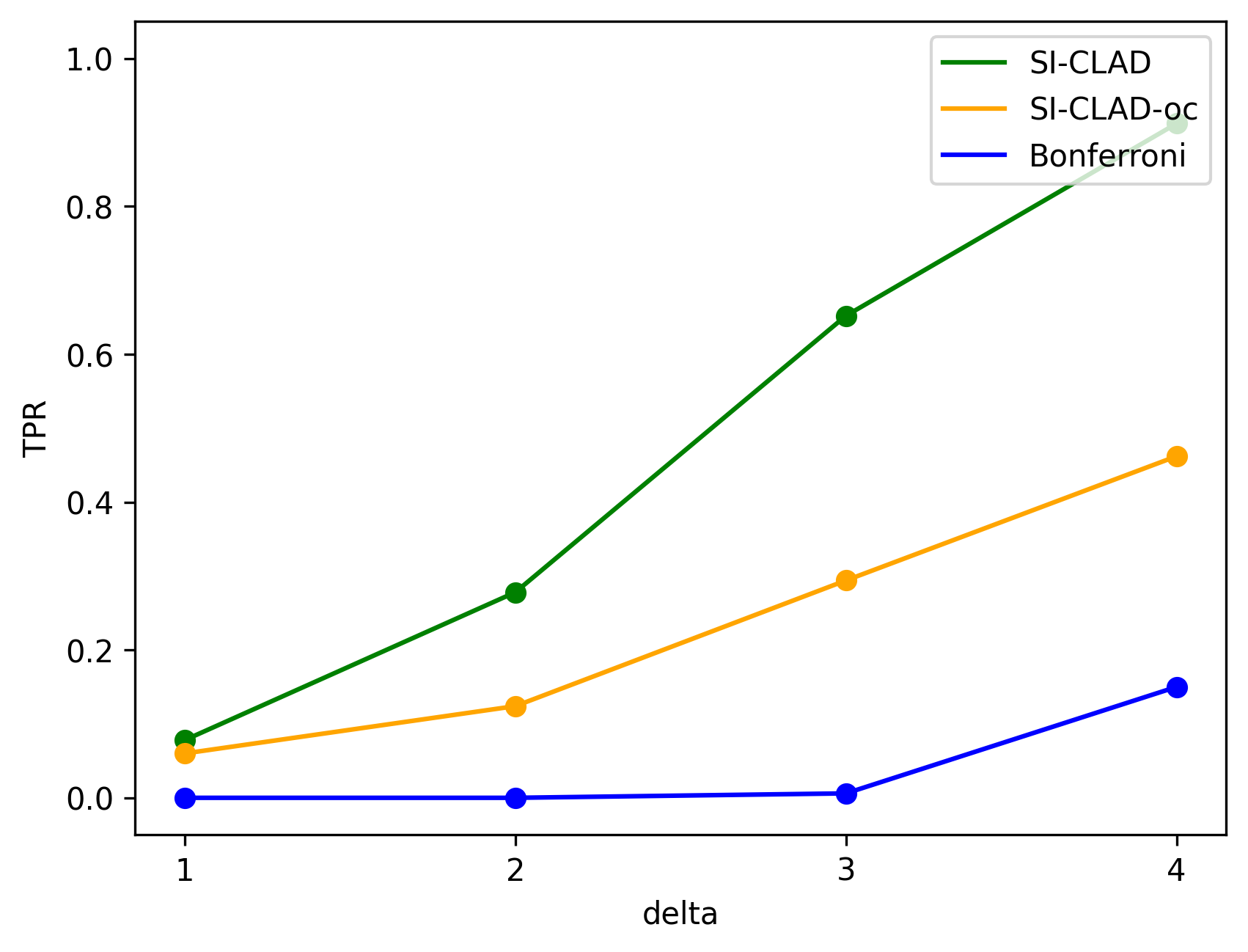} 
\end{subfigure}
\caption{FPR and TPR in multi-dimensional case} 
\label{fig:multiD}
\end{minipage}
\end{figure}

\textbf{Univariate case.} We generated $\bm X$ with $\mu_i = 0, \veps_i \sim \mathbb{N}(0,1)$, for all $i \in [n]$. We randomly selected $\left\lfloor\frac{n}{3} \right\rfloor$ data points and made them to be abnormal by setting $\mu_i = \mu_i + \Delta$. Regarding the FPR experiments, we set $n \in \{50,100,150,200\}$ and $\Delta = 0$. In regard to the TPR experiments, we set $n = 100$ and $\Delta = \{ 1,2,3,4\}$. Each experiment was repeated 500 times with $MinPts= 5, eps= 0.2$. The result are shown in Fig. \ref{fig:1d}. In the plot on the left, the {\tt SI-CLAD}, {\tt SI-CLAD-oc} and {\tt Bonferroni} controlled the FPR under $\alpha$ whereas the {\tt Naive} and {\tt No-Inference} could not. Because the {\tt Naive} and {\tt No-Inference} failed to control FPR, we no longer considered the TPR of those methods. In the plot on the right, we can see that the {\tt SI-CLAD} has highest TPR compared to other methods in all cases, i.e., {\tt SI-CLAD} has the lowest FNR.

%
%
%

\textbf{Multi-dimensional case.} We generated $\bm X$ with $X_{i,:} \sim \mathbb{N}(\bm 0_d, I_d), \forall i \in [n]$, the dimension $d=5$,~ $MinPts=10$, and $eps = 3$.
The settings of case FPR experiments were the same as univariate case. The result is shown in Fig. \ref{fig:multiD}. 
In multi-dimensional case, the proposed {\tt SI-CLAD} has highest TPR while controlling FPR under $\alpha$.

\begin{figure}[!t]
\begin{minipage}{.48\linewidth}
\begin{subfigure}{\linewidth}
  \centering
  \includegraphics[width=\linewidth]{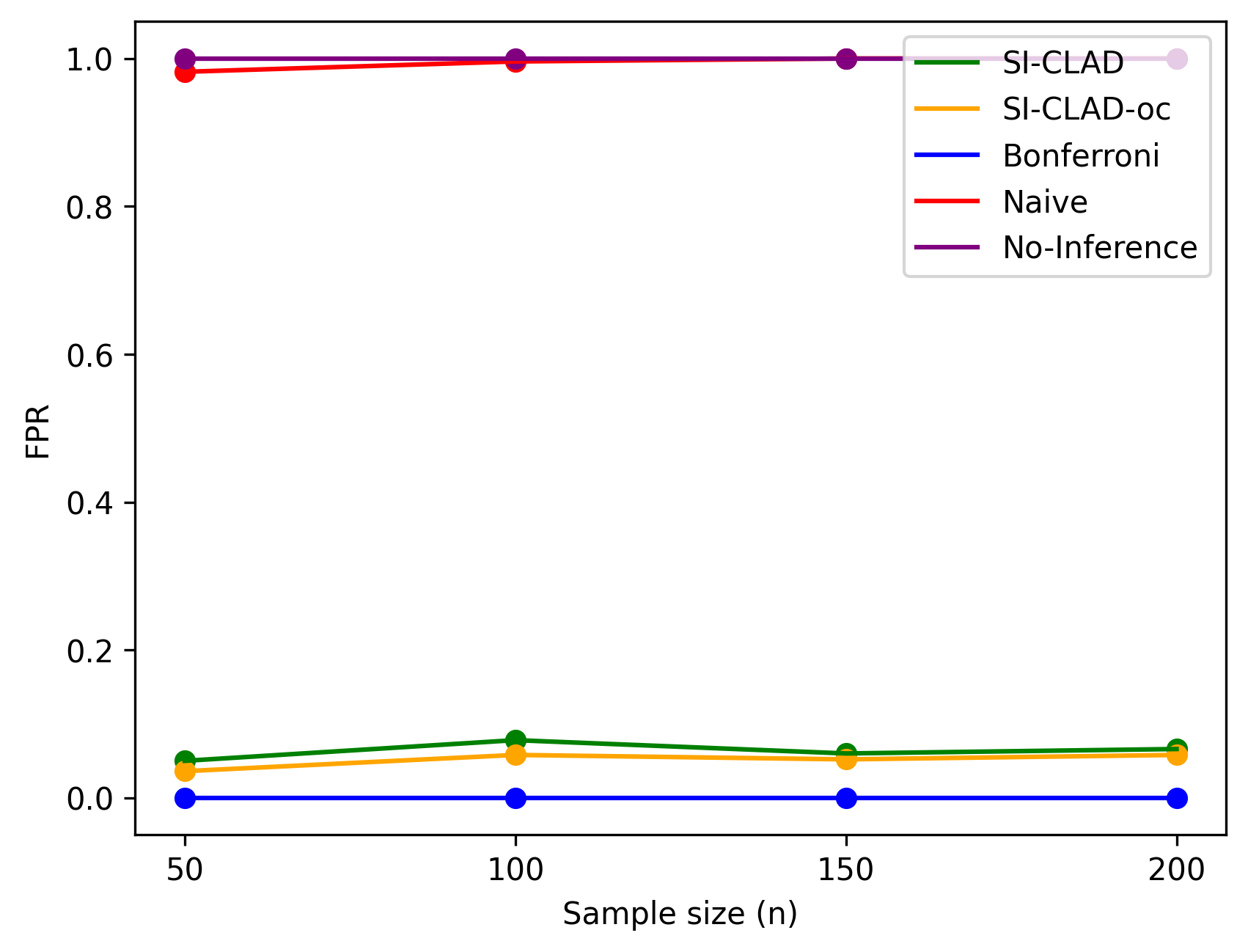}  
\end{subfigure}
\begin{subfigure}{\linewidth}
  \centering
  \includegraphics[width=\linewidth]{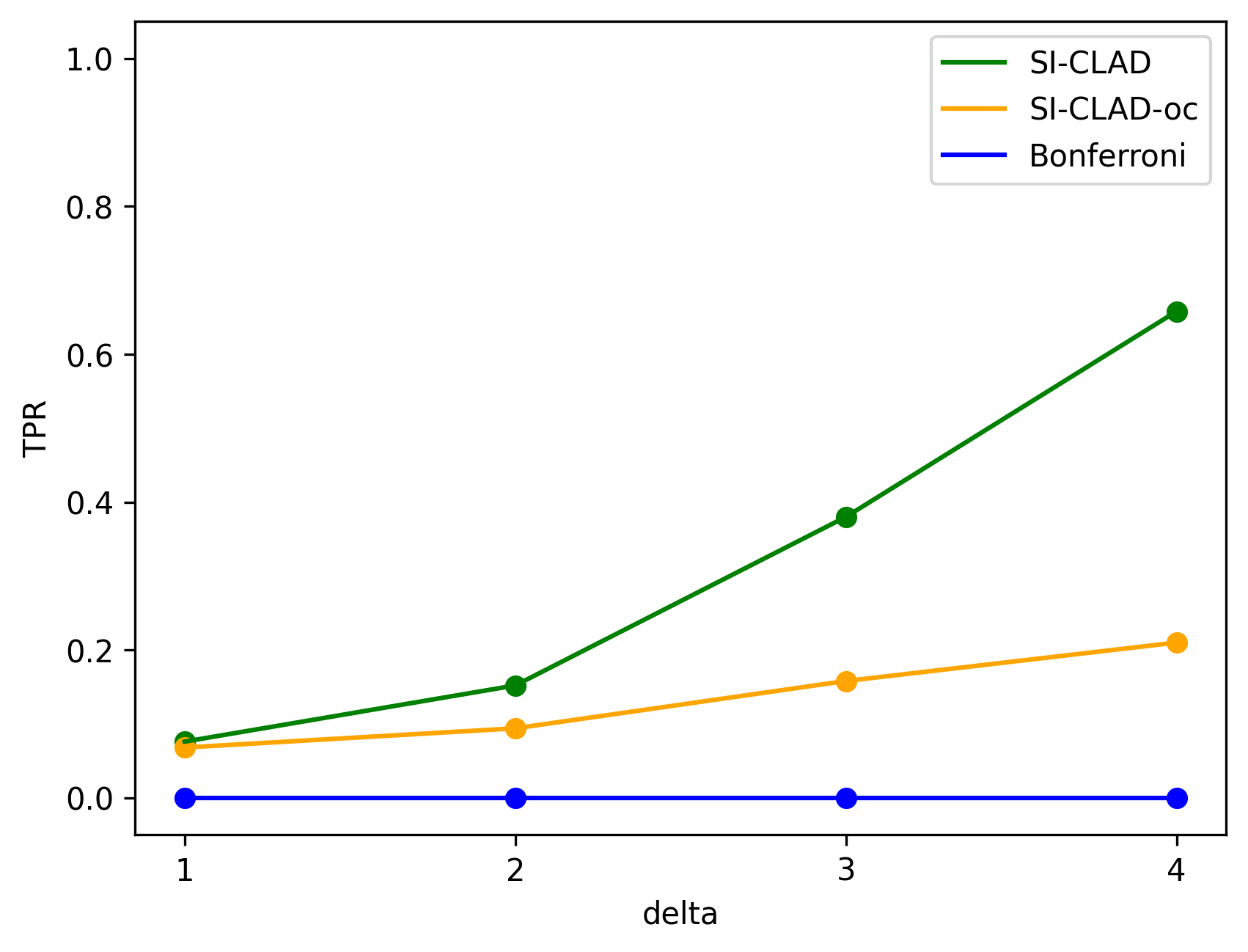} 
\end{subfigure}
\caption{FPR and TPR in the case of correlated data} 
\label{fig:corr}
\end{minipage}
\hspace{10pt}
\begin{minipage}{.48\linewidth}
\begin{subfigure}{\linewidth}
  \centering
  \includegraphics[width=\linewidth]{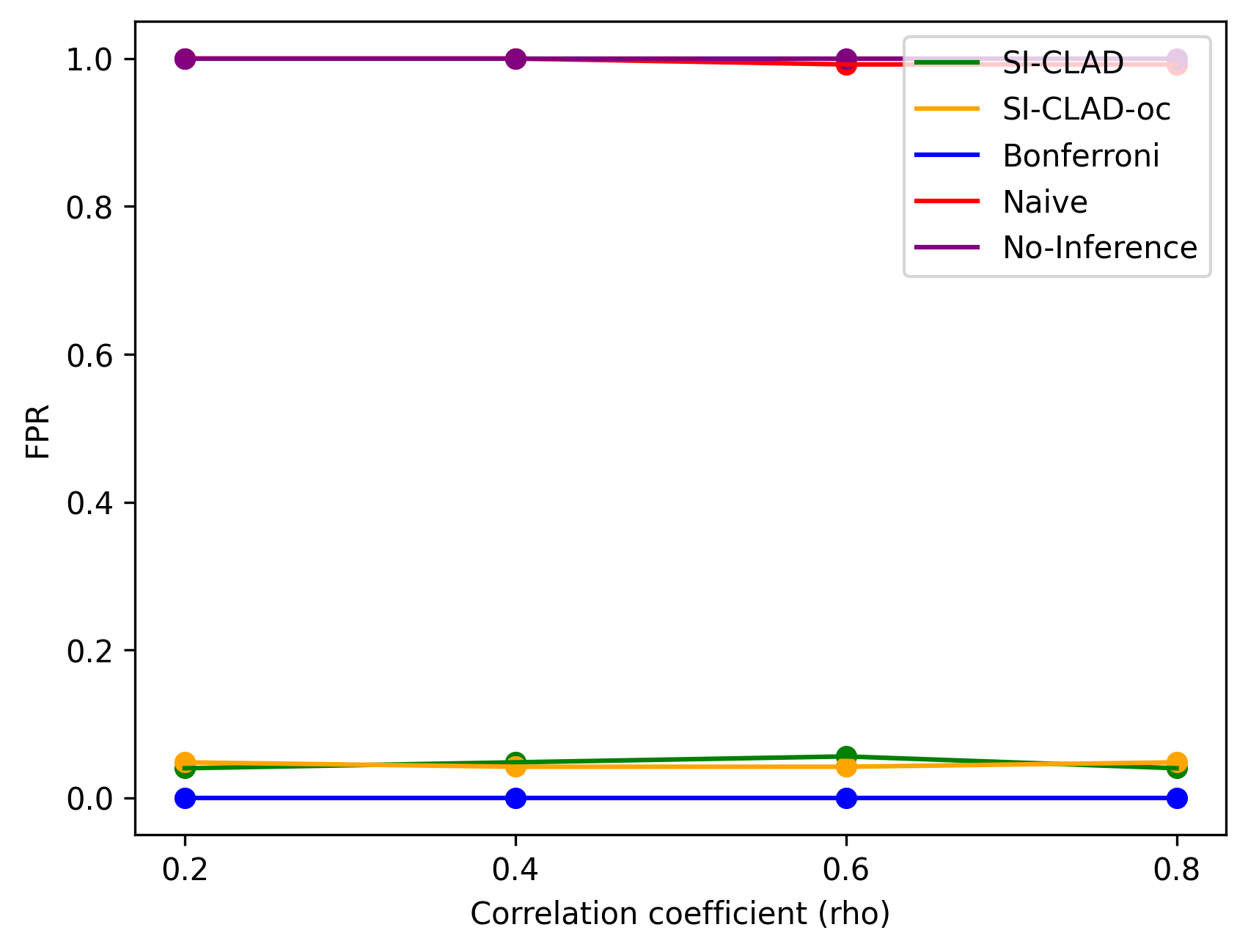}  
\end{subfigure}
\begin{subfigure}{\linewidth}
  \centering
  \includegraphics[width=\linewidth]{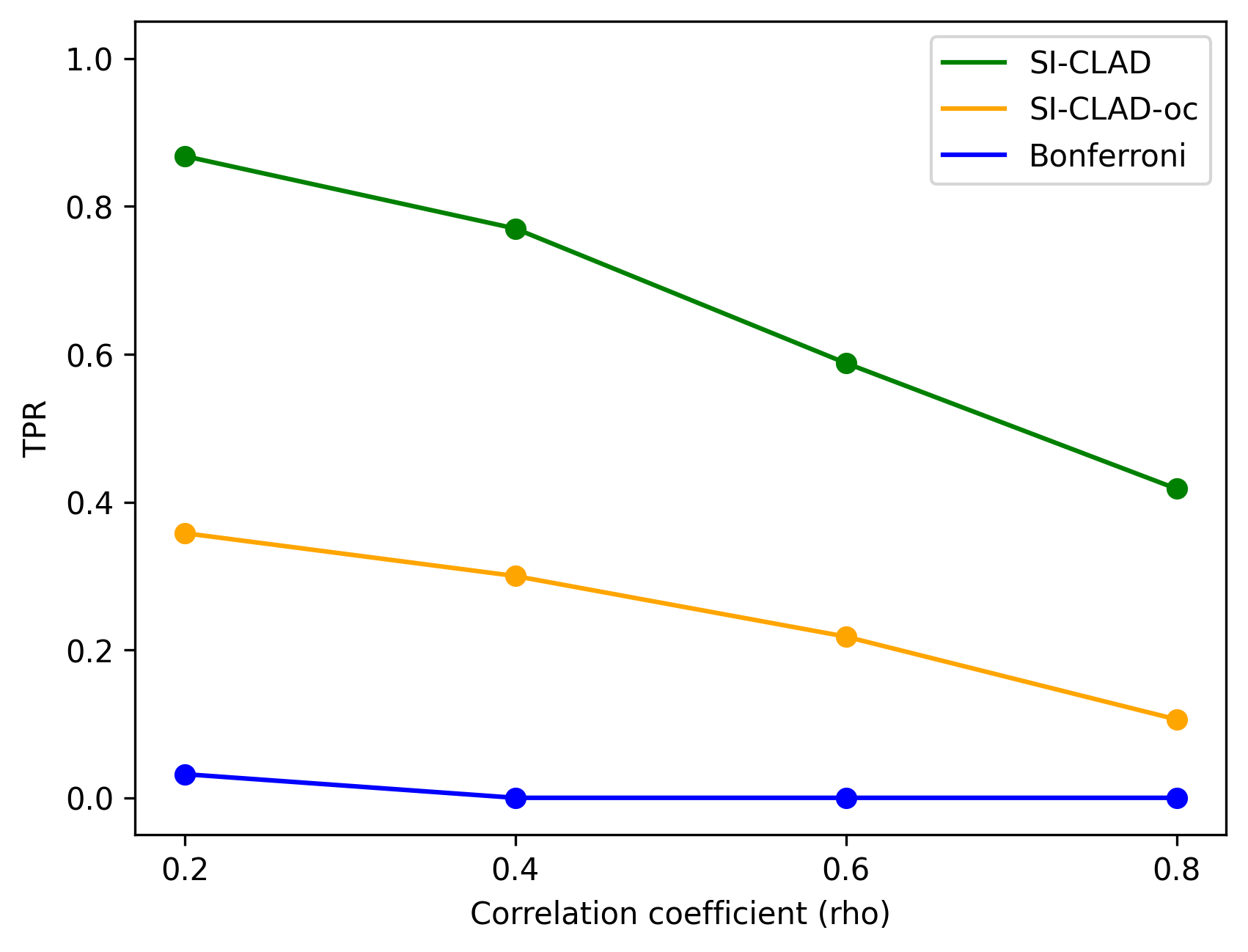} 
\end{subfigure}
\caption{FPR and TPR when changing $\rho$} 
\label{fig:corr_change_rho}
\end{minipage}
\end{figure}

\textbf{Correlated data.}  In this setting, we consider the case where the data is correlated. We generated $X$ with $X_{i,:} \sim \mathbb{N}(\bm 0_d, \Xi),~\forall i \in [N]$, the matrix $\Xi = \left[ \rho^{|i-j|}\right]_{ij},~\forall i,j \in [d],~\rho = 0.5$. The setting for FPR and TPR experiments were also the same as multi-dimensional case. The results are shown in Fig. \ref{fig:corr}. Additionally, we also conducted FPR and TPR experiments when changing $\rho \in \{0.2,0.4,0.6,0.8\}$. We set $n=100, ~d = 5, ~MinPts = 10,~ eps = 2$. For FPR experiments, $~\Delta = 0$ and for TPR experiments, $\Delta = 4$. The results are shown in Fig. \ref{fig:corr_change_rho}.
In essence, the correlated data contains redundant information. This means that the effective sample size is smaller than the actual sample size. A smaller effective sample size reduces the amount of information available for the statistical test, making it less powerful. Therefore, the TPR tends to decrease when increasing the value of $\rho$. 
However, in all the case, the proposed {\tt SI-CLAD} method consistently achieves the highest TPR while controlling the FPR.

%

\begin{figure}[!t]
\begin{minipage}{.48\linewidth}
\begin{subfigure}{\linewidth}
  \centering
  \includegraphics[width=\linewidth]{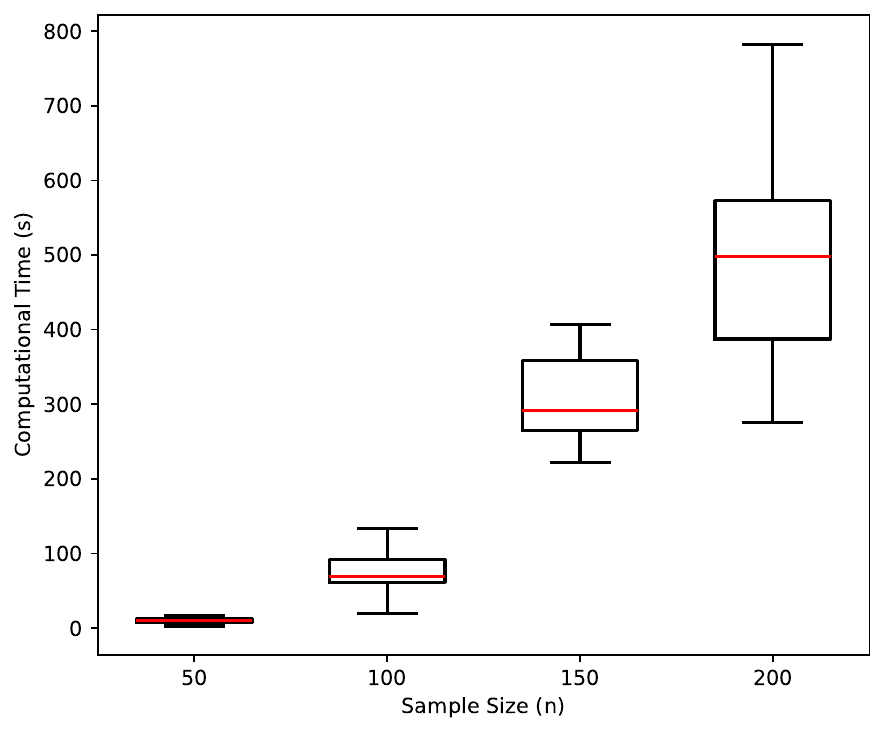}  
\end{subfigure}
\begin{subfigure}{\linewidth}
  \centering
  \includegraphics[width=\linewidth]{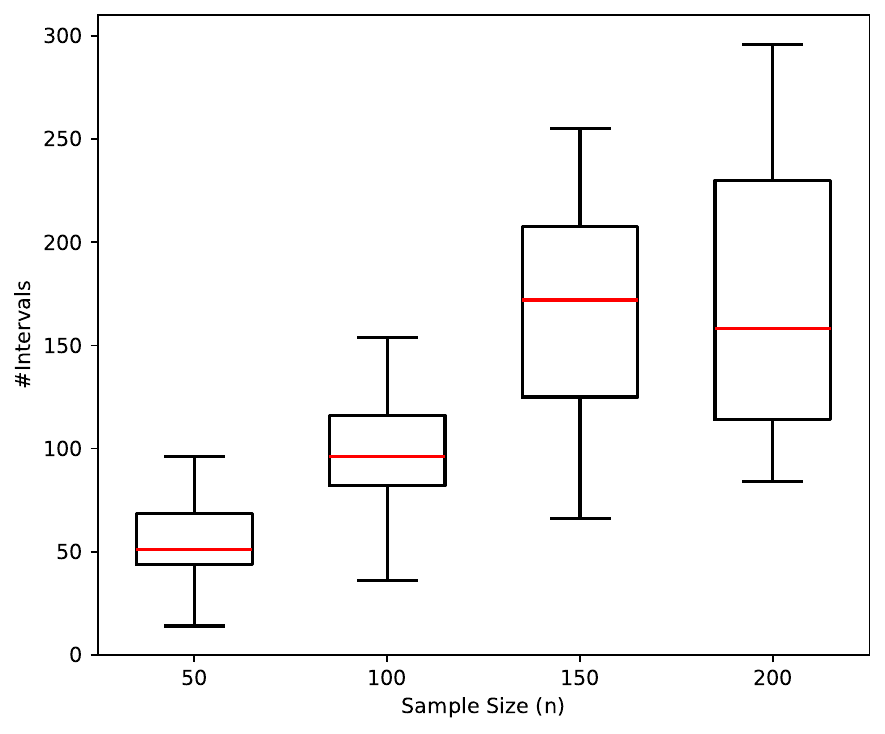} 
\end{subfigure}
\caption{Computational cost when changing $n$} 
\label{fig:cost_change_n}
\end{minipage}
\hspace{10pt}
\begin{minipage}{.48\linewidth}
\begin{subfigure}{\linewidth}
  \centering
  \includegraphics[width=\linewidth]{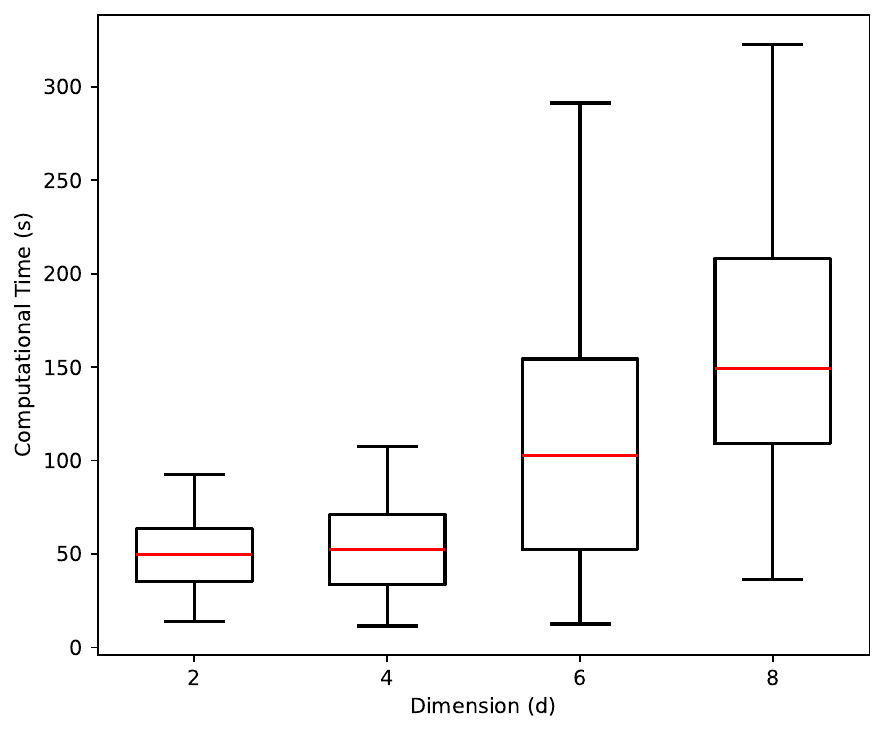}  
\end{subfigure}
\begin{subfigure}{\linewidth}
  \centering
  \includegraphics[width=\linewidth]{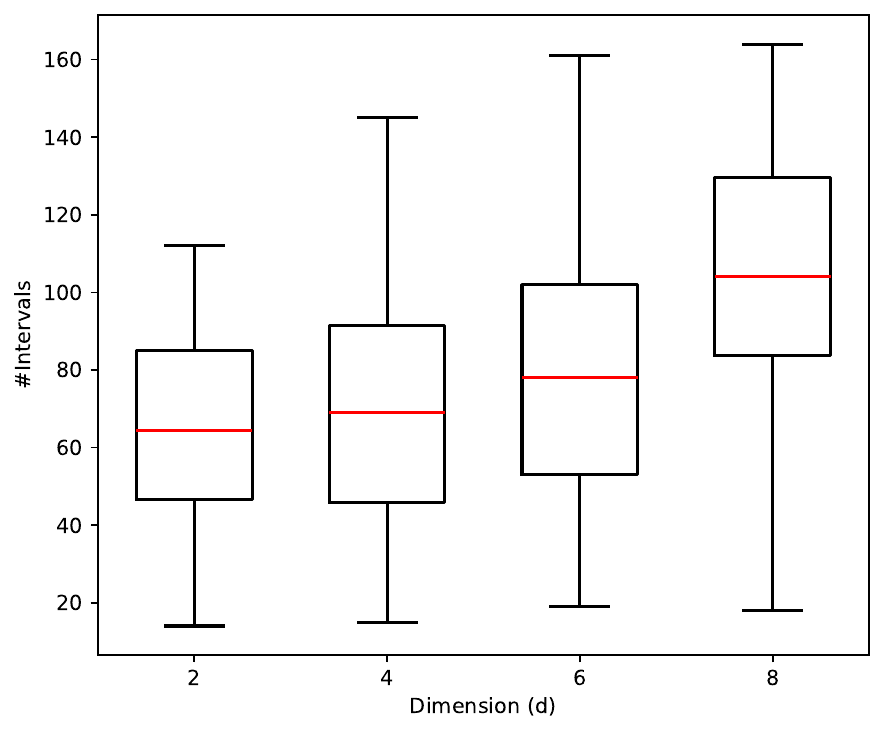} 
\end{subfigure}
\caption{Computational cost when changing $d$} 
\label{fig:cost_change_d}
\end{minipage}
\end{figure}

\textbf{Computational cost.}
Figure \ref{fig:cost_change_n} presents boxplots of the computational time per $p$-value and the number of intervals encountered during the construction of the truncation set $\cZ$ as functions of the sample size $n$. The results show that both computational time and the number of intervals increase approximately linearly with $n$, suggesting that the computational cost of {\tt SI-CLAD} scales efficiently with larger sample sizes.
Figure \ref{fig:cost_change_d} shows a similar analysis with respect to the dimension $d$. As $d$ increases, the computational time per $p$-value and the number of intervals also grow linearly.

 \subsection{Real-data Experiments}
\begin{figure}[!t]
    \centering
    \begin{subfigure}[b]{0.32\linewidth}
        \includegraphics[width=\linewidth]{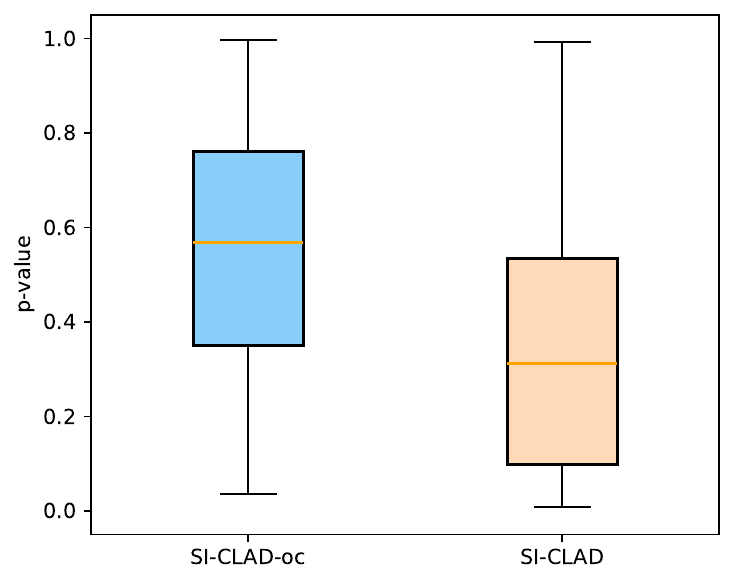}
        \caption{Heart Disease dataset}
        \label{fig:heartdisease}
    \end{subfigure}
    \hspace{0.02\linewidth}
    \begin{subfigure}[b]{0.32\linewidth}
        \includegraphics[width=\linewidth]{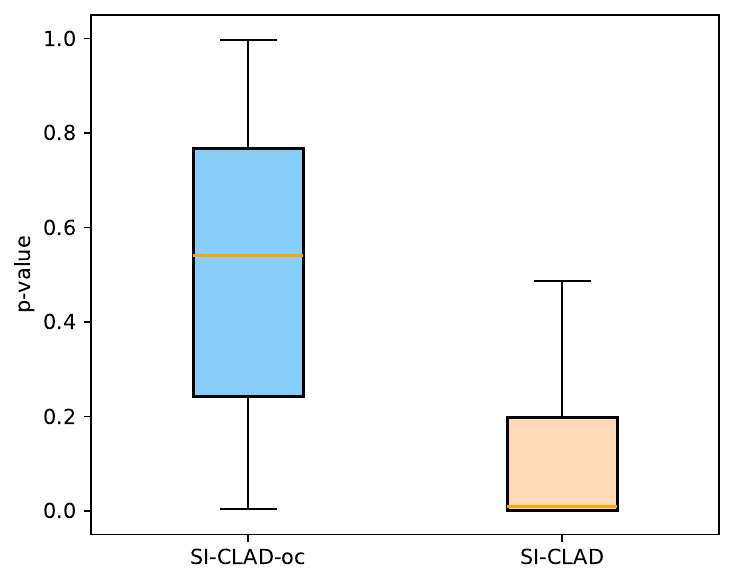}
        \caption{Breast Cancer dataset}
        \label{fig:breastcancer}
    \end{subfigure}
    \begin{subfigure}[b]{0.32\linewidth}
        \includegraphics[width=\linewidth]{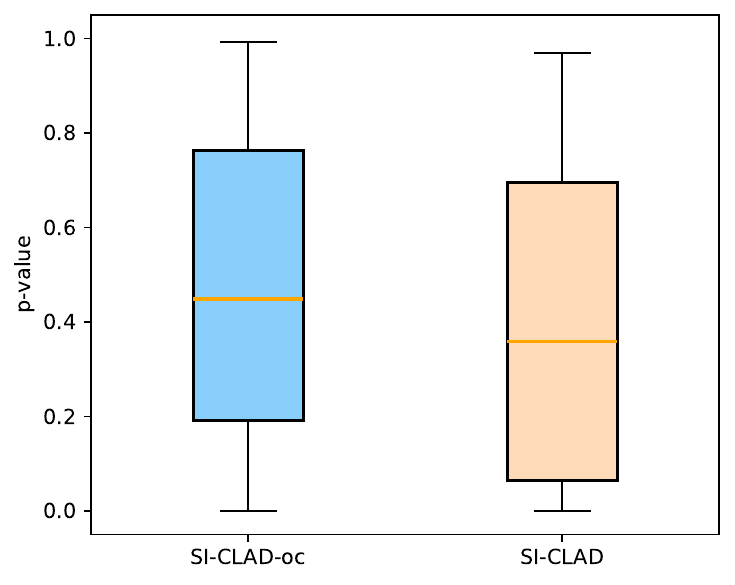}
        \caption{Absenteeism at work dataset}
        \label{fig:abs}
    \end{subfigure}
    \caption{Boxplots of $p$-values on real datasets}
    \label{fig:realdata}
\end{figure}

 We conducted comparisons on three real-world datasets: Breast Cancer Wisconsin, Heart Disease and Absenteeism at work available at the UCI Machine Learning Repository. We compared the $p$-values of the {\tt SI-CLAD} and {\tt SI-CLAD-oc}. 
 \begin{itemize}
     \item  \textbf{Heart Disease dataset} contains 303 patients with 13 features to predict heart disease presence. We randomly selected $n=200$ and set $MinPts = 26,~eps = 4$.
     \item \textbf{Breast Cancer Wisconsin (Diagnostic) dataset }has 569 instances with 30 features to classify tumors as benign or malignant. We randomly selected $d=15$ features, $n=200$ and set $MinPts = 30,~eps =5$.
     \item \textbf{Absenteeism at Work dataset} with a numerical target variable representing total employee's absenteeism time in a Brazilian company. For our unsupervised task, we treat all 20 variables - including the target - as regular features. We randomly select $n=200$ and set $MinPts = 40,~eps=6$.
 \end{itemize}
The box plots of the distribution of the $p$-values are illustrated in Fig. \ref{fig:realdata}. The p-values of the {\tt SI-CLAD} tend to be smaller than those of {\tt SI-CLAD-oc}, which indicates that the proposed {\tt SI-CLAD} method has higher power than the {\tt SI-CLAD-oc}.

%% file: sec5.tex
\section{Discussion}
\label{sec:sec5}
We propose a novel method to control the FPR under a given significance level $\alpha$, while maintaining a high TPR for statistical testing on anomalies detected by clustering-based AD.
Our approach leverages the Selective Inference (SI) framework and the parametric programming technique introduced in \cite{duy2021more} to efficiently compute the $p$-values for conducting the inference.
This study represents a significant advancement toward reliable artificial intelligence.
Some open questions remain.
While our current focus is on DBSCAN, the proposed statistical framework can be extended to test AD results obtained from other clustering techniques, such as OPTICS and DENCLUE.
Additionally, the hyper-parameters are currently provided by the analyst. If a data-driven procedure is used to determine these parameters, our method can also be extended to such settings—as long as the entire selection process can be characterized by linear or quadratic inequalities.

%% file: appendix.tex
\section{Appendix}
\label{app:A}
\subsection{Proof of Lemma \ref{lemma:valid_selective_p}}
We have 
\[
\bm \eta_j^\top \bm X \mid
	\left \{  
	\cO_{\bm X}
	=
	\cO_{\rm obs},
	\cQ_{\bm X}
	=
	\cQ_{\rm obs}
	\right \} 
	\sim 
	{\rm TN} 
	\left (
	\bm \eta_j^\top 
	{\bm \mu},
	\bm \eta_j^\top \Sigma \bm \eta_j,
	\cZ
	\right ),
\]
which is a truncated normal distribution with a mean $\bm \eta_j^\top {\bm \mu}$,
 variance $\bm \eta_j^\top \Sigma \bm \eta_j$ , and the truncation region $\cZ$ described in Sec. \ref{subsec:conditional_data_space}.
Therefore, under the null hypothesis,
\begin{align*}
	p_j^{\rm selective}
	\mid 
	\{ \cO_{\bm X}
	=
	\cO_{\rm obs},
	\cQ_{\bm X}
	=
	\cQ_{\rm obs}
	\}
	\sim {\rm Unif}(0, 1).
\end{align*} 
Thus,
$
	\mathbb{P}_{\rm H_{0, j}}  \Big (
	p_j^{\rm selective} \leq \alpha
	\mid 
	\cO_{\bm X}
	=
	\cO_{\rm obs},
	\cQ_{\bm X}
	=
	\cQ_{\rm obs}
	\Big) = \alpha, \forall \alpha \in [0, 1].
$

Next, we have 
\begin{align*}
	&\mathbb{P}_{\rm H_{0, j}}  \Big (p_j^{\rm selective} \leq \alpha \mid \cO_{\bm X} = \cO_{\rm obs} \Big ) \\ 
	&= 
	\int
	\mathbb{P}_{\rm H_{0, j}}  \Big (p_j^{\rm selective} \leq \alpha \mid \cO_{\bm X} = \cO_{\rm obs},  \cQ_{\bm X} = \cQ_{\rm obs} \Big ) ~
	\mathbb{P}_{\rm H_{0, j}}  \Big (\cQ_{\bm X} = \cQ_{\rm obs} \mid \cO_{\bm X} = \cO_{\rm obs} \Big ) d\cQ_{\rm obs} \\ 
	&= 
	\int \alpha 
	~ \mathbb{P}_{\rm H_{0, j}}  \Big (\cQ_{\bm X} = \cQ_{\rm obs} \mid \cO_{\bm X} = \cO_{\rm obs} \Big ) d\cQ_{\rm obs} \\ 
	& = 
	\alpha 
	\int \mathbb{P}_{\rm H_{0, j}}  \Big (\cQ_{\bm X} = \cQ_{\rm obs} \mid \cO_{\bm X} = \cO_{\rm obs} \Big ) d\cQ_{\rm obs} \\ 
	& = 
	\alpha.
\end{align*} 
Finally, we obtain the result in Lemma \ref{lemma:valid_selective_p} as follows:
\begin{align*}
	\mathbb{P}_{\rm H_{0, j}}  \Big (p_j^{\rm selective} \leq \alpha \Big ) 
	& = \sum \limits_{\cO_{\rm obs}}
	\mathbb{P}_{\rm H_{0, j}}  \Big (p_j^{\rm selective} \leq \alpha \mid \cO_{\bm X} = \cO_{\rm obs} \Big ) ~
	\mathbb{P}_{\rm H_{0, j}}  \Big (\cO_{\bm X} = \cO_{\rm obs} \Big ) \\ 
	& = \sum \limits_{\cO_{\rm obs}} \alpha ~ \mathbb{P}_{\rm H_{0, j}}  \Big (\cO_{\bm X} = \cO_{\rm obs} \Big ) \\ 
	& = \alpha \sum \limits_{\cO_{\rm obs}} \mathbb{P}_{\rm H_{0, j}}  \Big (\cO_{\bm X} = \cO_{\rm obs} \Big ) \\ 
	& = \alpha.
\end{align*}

\subsection{Proof of Lemma \ref{lemma:data_line}}
According to the second condition in \eq{eq:conditional_data_space}, we have 
\begin{align*}
	\cQ_{\bm X} & =  \cQ_{\rm obs} \\ 
	\Leftrightarrow 
	\left ( 
	I_{n} - 
	\bm b
	\bm \eta_j^\top \right ) 
	\bm X
	& = 
	\cQ_{\rm obs}\\ 
	\Leftrightarrow 
	\bm X
	& = 
	\cQ_{\rm obs}
	+ \bm b
	\bm \eta_j^\top  
	\bm X.
\end{align*}
By defining 
$\bm a = \cQ_{\rm obs}$,
$z = \bm \eta_j^\top \bm X$, and incorporating the first condition of \eq{eq:conditional_data_space}, we obtain Lemma \ref{lemma:data_line}. 

\subsection{Proof of Lemma \ref{lemma:over_conditioning}}
Define $\mathcal{I} = \{(i, j) : i \in [n], \, j \in [n]\}$ and 
$\sigma_{ij} = 
\begin{cases} 
+1 & \text{if } j \in N_{\varepsilon}(X_i'), \\ 
-1 & \text{if } j \notin N_{\varepsilon}(X_i'), 
\end{cases}$. The over-conditioned region \(\mathcal{Z}^{oc}\big (\bm X^\prime \big )\) defined in Equation \eqref{eq:z_oc} can be rewriten as:
\[
\mathcal{Z}^{oc}\big (\bm X^\prime \big ) =  
\bigcap_{(i, j) \in \mathcal{I}} 
\left\{ z \in \mathbb{R} \,\middle|\, 
\sigma_{ij} \|\bm X _i(z) - \bm X _j(z)\|^2 \leq eps^2 
\right\}
\]

\noindent
Let  \( \bm e_j \)  be a vector in \( \mathbb{R}^n \) with the \( j^{\mathrm{th}} \) entry equal to \( 1 \) and all other entries equal to \( 0 \). The condition $\sigma_{ij}\|\bm X^\prime _i(z) - \bm X^\prime _j(z)\|^2 \leq eps^2$ can be expressed as: 
\[
\begin{aligned}
\sigma_{ij}\left\|\bm{e}^\top_i \bm X (z)- \bm{e}^\top_j \bm X(z)\right\|^2 &\leq eps^2\\ 
\Leftrightarrow \sigma_{ij}\bm X (z)^\top (\bm{e}^\top_i - \bm{e}^\top_j)^\top (\bm{e}^\top_i - \bm{e}^\top_j) \bm X (z) &\leq eps^2\\
\Leftrightarrow \sigma_{ij}\bm X (z)^\top (\bm{e}_i - \bm{e}_j) (\bm{e}_i - \bm{e}_j)^\top \bm X (z) &\leq eps^2.
\end{aligned}
\]
For simplicity of notation, define \(\bm{A}_{ij} = (\bm{e}_i - \bm{e}_j) (\bm{e}_i - \bm{e}_j)^\top\). Substituting the parameterired form of \(\bm X (z)\) into the inequality:  
\[
\begin{aligned}
\sigma_{ij}(\bm{a} + \bm{b}z)^\top \bm{A}_{ij} (\bm{a} + \bm{b}z) &\leq eps^2\\ 
\Leftrightarrow (\sigma_{ij}\bm{b}^\top \bm{A}_{ij} \bm{b})z^2 + (\sigma_{ij}\bm{b}^\top \bm{A}_{ij} \bm{a} + \sigma_{ij}\bm{a}^\top \bm{A}_{ij} \bm{b})z + (\sigma_{ij}\bm{a}^\top \bm{A}_{ij} \bm{a} - eps^2) &\leq 0,
\end{aligned}
\]  
Let us define \[
\bm p = (\bm p_{ij})_{(i,j) \in \cI}, \quad \bm q = (\bm q_{ij})_{(i,j) \in \cI}, \quad \bm t = (\bm t_{ij})_{(i,j) \in \cI} ~,
\]
where
\[
   \bm p_{ij} = \sigma_{ij}\bm{b}^\top \bm{A}_{ij} \bm{b}, \quad\bm q_{ij} =\sigma_{ij} \bm{b}^\top \bm{A}_{ij} \bm{a} + \sigma_{ij}\bm{a}^\top \bm{A}_{ij} \bm{b}, \quad \bm t_{ij} = \sigma_{ij}\bm{a}^\top \bm{A}_{ij} \bm{a} - eps^2.\]
Substituting them into the inequality, we obtain:  
\[\cZ^{\rm oc}\big (\bm X^\prime \big ) =  \left\{ z \in \mathbb{R} \;\middle|\; \bm p z^2 + \bm q z + \bm t \leq 0 \right\}\]
This completes the proof.

While various distance functions can be applied to query the 
$eps$-neighborhood in DBSCAN, it is important to note that the set $R_{ij}$ above is specifically defined using the Euclidean distance metric. This means that our proposed method is designed to inference the anomalies from the DBSCAN algorithm that uses this distance function.